\documentclass{article}
\usepackage{ismir,amsmath,cite,url}
\usepackage{graphicx}
\usepackage{color}

\usepackage{microtype}
\usepackage{amsmath}
\usepackage{cases}
\usepackage{subfigure}
\usepackage{filecontents}
\usepackage{pgfplots}
\usepackage{booktabs} 
\usepackage{lilyglyphs}
\usepackage{musixtex}
\usepackage{CJK}
\usepackage{{graphicx}}

\title{Melodic Phrase Segmentation By Deep Neural Networks}

\multauthor
{Yixing Guan \textsuperscript{*} \hspace{1cm} Jinyu Zhao \textsuperscript{*} \thanks{* Equal Contribution} \hspace{1cm} Yiqin Qiu} {\bfseries{Zheng Zhang \hspace{1cm} Gus Xia}\\
New York University Shanghai\\
{\tt\small {\{yixing.guan, jinyu.zhao, yiqin.qiu, zz, gxia}\}@nyu.edu}
}
\def\authorname{Yixing Guan, Jinyu Zhao, Yiqin Qiu, Zheng Zhang, Gus Xia}

\begin{document}
\maketitle
\begin{abstract}
Automated melodic phrase detection and segmentation is a classical task in content-based music information retrieval and also the key towards automated music structure analysis. However, traditional methods still cannot satisfy practical requirements. In this paper, we explore and adapt various neural network architectures to see if they can be generalized to work with the symbolic representation of music and produce satisfactory melodic phrase segmentation. The main issue of applying deep-learning methods to phrase detection is the \textit{sparse labeling} problem of training sets. We proposed two tailored label engineering with corresponding training techniques for different neural networks in order to make decisions at a sequential level. Experiment results show that the CNN-CRF architecture performs the best, being able to offer finer segmentation and faster to train, while CNN, Bi-LSTM-CNN and Bi-LSTM-CRF are acceptable alternatives.
\end{abstract}

\section{Introduction} \label{sec:introduction}

Automated melodic phrase detection and segmentation is a classical task in content-based music information retrieval (MIR). It is also the key step towards automated music structure analysis \cite{MIREX2009}, which is useful for many computer-music applications, such as structured automated composition \cite{nierhaus2009algorithmic} , music databases \cite{PEDB}, and query-by-humming \cite{ghias1995query}. However, current solutions for melodic phrase detection cannot yet satisfy practical requirements, especially for symbolic music representation. To be specific, rule-based methods in general rely on theme repetitions, long notes and rests, and hence are unstable when dealing with music with large variations; traditional machine-learning methods rely on manually-designed features and very difficult to capture useful music context information for boundary detection. 

On the other hand, many neural network architectures have recently achieved quite promising results in various domains, including representation learning\cite{representation_learning}, computer vision \cite{Computer_Vision}, natural language processing \cite{NLP}, autonomous driving \cite{Autonomous_Driving}. Since one can naturally consider a piece of melody as an array of notes with given pitches and lengths (ordered by their onsets), we set out to automatically merge consecutive notes into a larger unit and this representation of music spontaneously turns our objective into a supervised learning problem, which allows the use of some existing neural network architectures.

Analogous to adding punctuation to an array of characters to form sentences for NLP application, in this paper, we are primarily interested in labeling the begin and end of a phrase given an array of notes, which then gives us segmentation masks of phrases, if we deem each phrase distinct. Since the supervised neural network can use the back-propagation mechanism to automatically identify the crucial music-context related features in various ways, we use a combination of existing neural networks and probabilistic graphical models to solve the task, including CNN, Bi-LSTM, CNN-CRF, Bi-LSTM-CRF. 

The main issue of applying deep-learning methods to phrase detection is the \textit{sparse labeling} of the training sets. To address this issue, we:\\
1) contribute two label engineering techniques to solve the sparse labeling problem that hinders the use of sequential decision-making neural networks\\
2) combine the label engineering techniques with proper neural networks which considers both implicit and explicit relationships between labels to detect phrase boundaries in symbolic representations of music\\
3) conducted a quantitative evaluation of the performance of the proposed models for the task\\
All models are trained and tested on a customized dataset we collected. Experiment results show that the CNN-CRF architecture performs the best, being able to offer finer segmentation and faster to train, while CNN, Bi-LSTM-CNN and Bi-LSTM-CRF are acceptable alternatives. 

In the following sections, we discuss the related work in section \ref{sec:related_work} and introduce the formal problem definition in section \ref{sec:problem_definition}. We present the methodology in section \ref{sec:methodology} followed by the experimental results in section \ref{sec:experiments}. Finally, we conclude our paper with some reflections and possible directions for future works in section \ref{sec:conclusion}.

\section{RELATED WORK} \label{sec:related_work}

Many previous works on analyzing music structure are based on audio representation. Foote \cite{Self_Similarity} first proposed a method which visualizes the self-similarity between two instants in music given its audio input, which produces a 2D self-similarity matrix that can characterize the structure. Later, by measuring change in local self-similarity, Foote \cite{Self_Similarity_Segment} developed a classic method for automatic audio segmentation. Based on the notion of self-similarity, Kaiser and Sikora \cite{Non-negative} further proposed a method that applies non-negative matrix factorization to self-similarity matrix and produces two factorization products, upon which the structure boundary can be derived. Other methods include Hidden Markov Model \cite{HMM}, decision tree\cite{supervised2007} and clustering \cite{Cluster}. 

One very recent work of Ullrich et al \cite{CNN_Seg} adapts Convolutional Neural Networks to produce musical structure segmentation, which is most relevant to our work: their network takes a spectrogram as input and outputs the probability of a phrase probability of each spectrum, following by which peak-picking and thresholding are applied to post-process the result. Our study also considers the problem of phrase segmentation. Different from traditional methods based on symbolic representation which heavily rely on long music notes and rest \cite{symbolic}, our system is purely learning based and takes into account more music context. Moreover, rather than using a rule-based post-processing method as in \cite{CNN_Seg}, we used CRF and combined with the deep learning architecture, making the system end-to-end. The method of using CNN only is treated as the baseline method to be compared against in our study.

\section{PROBLEM DEFINITION} \label{sec:problem_definition}

In this section, we formally define our problem and introduce our data representation in detail. We denote \textbf{X} as the random variable over music sequence to be labeled and \textbf{Y} is a random variable over the space of all valid label sequences. A specific music phrase is denoted as $\{x_i\} = x \sim F_X$ , and $\hat{y_i} = \phi(x_i)$ is the predicted label generated by the model $\phi$. We use $y_i*$ as the corresponding ground truth label. Our goal is to construct the conditional probability $P(Y\vert X)$, which is approximated by $p(y|x)$ = $\phi_\theta(x)$, (x,y samples from the dataset $\mathcal{D}$). The model $\phi_\theta(x)$ is optimized using maximum likelihood estimation by finding $\underset{\theta \in \Omega} {\operatorname{argmax}}\ \mathcal{L}(y*,\phi_\theta(x))$ . While in practice we approach the problem by performing empirical risk minimization $\underset{\theta \in \Omega} {\operatorname{argmin}}\ \frac{1}{N} \sum_{i=1}^{N} Loss(y_i*,\phi_\theta(x_i))$ , where $\Omega$ is the parameter space.

\subsection{Music Phrase Representation}
Each music phrase contains multiple notes. We represent each note by its pitch, duration and offset-onset interval to the next note. This representation contains enough information to reconstruct the original music melody and encodes the time dimension within the sequence. All the methods we present in Section \ref{sec:methodology} use the same phrase representation.

\subsection{Label Representation}

Almost all previous works posed the musical phrase segmentation task naturally as a binary classification problem and thus deployed the binary labeling scheme. Consequently, dataset has highly-imbalanced label distribution, which makes the training process much harder for many neural network architectures. To solve the sparse labeling problem, we propose two alternative label engineering techniques, while keeping the 0-1 labeled dataset for the training of our baseline method.
One is named as "exponential-decay label", where we assign value 1 to the start of a music phrase, starting from where the value decays exponentially to ½, ¼ till it reaches to the middle of a sentence before it goes up with the same rate till the start of next phrase; the other is named as "linear-ascend label", where we assign to a note the value of that note's numbered position of the phrase it is in. Notes not in any phrases will be assigned label 0. 

A visualization of these three types of labeling is provided in Figure 1. The original binary labels are displayed in the first row, while the "exponential-decay label" and "linear-ascend label" are shown in the second and third row, respectively. This two labeling clearly solve the label imbalance problem, and in particular, the linear ascending label can be used to train neural network with CRF as its final layer, since the state transitional matrix of CRF requires a discrete label space.


\begin{figure}
	\includegraphics[width=\linewidth]{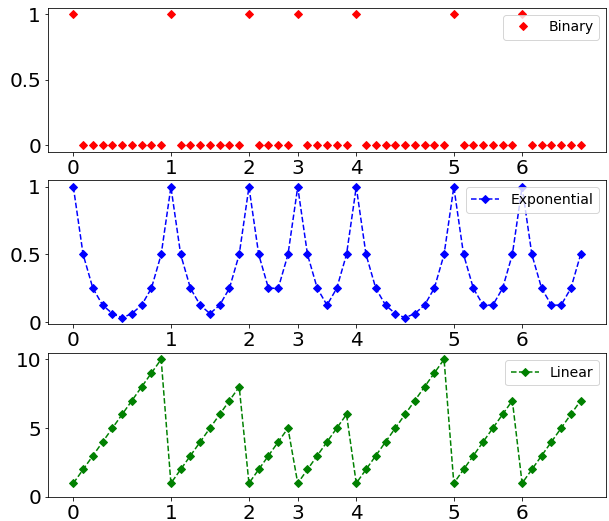}
    \caption{label representation comparison}
\end{figure}

\section{Methodology} \label{sec:methodology}
We present in section \ref{subsec:CNN} to \ref{subsec:CRF} three methods to tackle the automated phrase segmentation task, with a brief introduction to the neural network modules used in each method, while the loss functions used for training can be found in section 4.4.

\subsection{CNN Variants} \label{subsec:CNN}

Convolutional neural networks are mainly used in computer vision tasks such as object detecting and image segmentation \cite{Computer_Vision,unet}. The most important feature of convolution layers is that they are able to catch certain local features in their receptive fields while preserving most of the information on positioning. A typical CNN architecture consists of alternating convolution layers and pooling layers. 

For a convolution layer with input $X^{l-1}_{1..C_{l-1}}$, output $X^{l}_{1..C_{l}}$, weight $W$ and bias $b$, it computes as:
\begin{equation*}
X^{l}_{j} = \sum_{i} (b_{j}+X^{l-1}_i*W_{ij})
\end{equation*}
Where the sign * denotes the convolution, $0\leq i \leq C_{l-1}$ and $0\leq j \leq C_{l}$

Similar to how image segmentation is often performed on pixels of images, we try to use CNN to perform phrase segmentation on notes of the music.

\subsubsection{Single-direction CNN}
We implement a 5-layer CNN with a softmax layer on the top of the network. Note that different from usual CNN kernels whose receptive fields cover both sides of the neuron, we use kernels that only receive signals from the neurons in front of them, which is inspired from the architecture design of WaveNet \cite{wavenet}. This modification can increase the receptive field of the final-layer neurons so it can be useful in our target. We treat this modified version of CNN as our baseline method to be compared against, since \cite{CNN_Seg} has already shown that CNN can be used for phrase segmentation task.

\subsubsection{U-net}
Taking into consideration U-net's high performance on image segmentation tasks \cite{unet}, we adapt part of its architecture to put to use for our phrase segmentation task. To be more specific, a 5-layer depth top-down then bottom-up U-net architecture is implemented and used as one of the CNN variants. Each block consists of one convolution layer and one pooling/up-sample layer, with skip connections connecting blocks at each level of the network.

\subsection{LSTM Variants} \label{subsec:LSTM}

Long Short-Term Memory, usually referred as LSTM, falls into the category of Recurrent neural network, which aims to model correlation within sequence of data. Introduced first by \cite{LSTM}, given a sequence of data ${X_n}$, a LSTM unit typically consists of a memory cell $c$, an input gate $i$, a forget gate $f$ and an output gate $o$. The update formula for this LSTM unit at time $t$ can then be written as the following: 
\begin{align*}
\hat{c_t} &= tanh(U_c h_{t-1}+b_{U_c}+W_c x_t+b_{W_c})\\
i_t &= \sigma(U_f h_{t-1}+b_{U_i}+W_f x_t+b_{W_i})\\
f_t &= \sigma(U_i h_{t-1}+b_{U_f}+W_i x_t+b_{W_f})\\
o_t &= \sigma(U_o h_{t-1}+b_{U_o}+W_o x_t+b_{W_o})\\
c_t &= f_t\odot c_{t-1}+i_t\odot \hat{c_t}\\
h_t &= o_t\odot tanh(c_t)
\end{align*}
Here $W_c, W_i, W_f, W_o$ and $b_{W_c}, b_{W_i}, b_{W_f}, b_{W_o}$ denote the weight matrices and biases for the memory cell $c$, input gate $i$,  forget gate $f$ and output gate $O$ respectively. $h_t$ is the hidden state, as well as the output of the LSTM cell at time $t$. $\sigma$, $\odot$ are the element-wise sigmoid and product operators. 

In this paper, we use Bi-LSTM, the bi-directional LSTM module introduced first by \cite{BiLSTM} to build our neural network. The bi-directional LSTM(Bi-LSTM) runs the sequential input twice, forward and backward, to produce two hidden states, one of which encodes information from the past input, while the other encapsulates information from the future input. Concatenating these two hidden states give the final output. Since future music rhythms can indeed be informative, we prefer Bi-LSTM over the basic LSTM unit in this paper.

\subsubsection{Bi-directional LSTM with CNN}

While the single directional LSTM model mimics human pattern of listening to the music, it fails to leverage all the information encoded within a sequence of music. The one-way direction only enables the information flow from the past to present, without knowing what will happen next. The bi-directional LSTM solves the problem. In addition, it is also quite compatible with CNN layers as feature extractor. By stacking these two model together, we combines the advantages of them. Convolution kernel extracts local feature efficiently and the LSTM is capable to make temporal modeling of sequential feature \cite{CNN-LSTN-DNN}. Our architecture is motivated by \cite{BiLSTMCRF}, in which the paper uses CNN to compute the character-level representation and \cite{RNN_representation}, which indicates the better representation of the input feature could improve the performance of LSTM. Therefore, we combine CNN and LSTM into an unified framework and train them jointly. 

\subsection{CRF Variants} \label{subsec:CRF}

For sequential prediction problem, it's beneficial to take the correlation between labels into consideration. For example in our case, since it's really rare that a music phrase only consist of one note, the probability that two adjacent labels are both positive is close to 0. Therefore, Conditional Random Field, usually referred as CRF, can prevent such scenario from happening and jointly decode a sequence of labels that best pairs with the input data.

Formally speaking, let \textbf{X} be the random variable over the data sequences that we want to label, and we use \textbf{x} = \{$x_1$, $...$, $x_n$\} to represent a generic input sequence where $x_i$ is the $i_th$ pitch of the sentence. \textbf{y} = \{$y_1$,...,$y_n$\} represents the generic label sequences corresponding to \textbf{x}. The random variable \textbf{X} and \textbf{Y} are jointly distributed, yet CRF tries to model the conditional probability $P(\textbf{Y}|\textbf{X})$ from the observation pair ($x_i$, $y_i$) \cite{CRF}. The general CRF describes correlations between vertexes in any graph, while in our context, we focus on the special case where the graph \textbf{G} is a chain. Therefore, it reduces to the simple HMM-like CRF except in CRF all the $\{x_i\}$ are modeled together. Hence, we can factorize our graph and define potential functions dependent on \textbf{x}, $y_i$ and $y_{i-1}$. By fundamental theorem of random graph\cite{randomgraph}, the log conditional probability $log P(\textbf{Y}|\textbf{X})$ is in direct proportion to the linear combination of score functions $f_k(e,y|_e, \textbf{x})$ and $g_k(v,y|_v, \textbf{y})$ where $y|_S$ is the set of components associate to the subgraph $S$. As a result, the probabilistic model for sequence given  \textbf{x} and \textbf{y} can be computed by the following form:
\begin{equation*}
	P(y|x) =  \frac{\prod_{i=1}^{n} \varphi_{i}(y_{i-1}, y_i, \textbf{x})}{\sum_{y' \in \textit{L}(\textbf{x})} \prod_{i=1}^{n} \varphi_{i}(y'_{-1}, y'_i, \textbf{x})}
\end{equation*}
where \textit{L}(\textbf{x}) denotes the label space with size m and $\varphi_{i}(y', y, \textbf{x}) = exp(\textbf{W}_{y',y}^{T} x_i + b_{y', y})$. In practice, we simplified it to $\varphi_{i}(y_{i-1}, y_i, \textbf{x}) = exp(\psi_{emit}(y_i \mapsto x_i) + \psi_{trans}(y_{i-1} \mapsto y_i)) =h_i[y_i] + P_{y_i, y_{i-1}}$, where $h_i$ is a 1 by m matrix computed by bottom architecture and $P_{y',y}$ is the state transition matrix.
 
Inspired by \cite{CRFRNN}, in this paper, we also try to treat CNN and LSTM as merely feature extractors and our objective as pure sequence labeling problem, which produces results beyond our expectation.

\subsubsection{CNN-CRF}

This CRF variant has a 7-layer CNN with skip-connection and ReLU activation function as its feature extractor. The kernel size is 3 for all CNN layers. Again, we did not claim 7 as an optimal number for the task. The intuition is that we use CNN to directly model the inter-dependency among the notes within the receptive field of the multi-layer CNN feature extractors, since LSTM model can only theoretically model all forms of inter-dependencies within a sequence.

\subsubsection{Bi-LSTM-CRF}

Inspired by the use of Bi-LSTM-CRF for sequence tagging problem \cite{BiLSTMCRF}, we decide to give this architecture a shot as well. We use the same Bi-LSTM module mentioned in section \ref{subsec:LSTM} as the bottom layer, upon which we build our linear-chain CRF.

\subsection{Loss Functions}

Since we have three methods, each of which can be paired with different labels to train for the automated phrase segmentation task, we introduce here in details the loss functions we used in the training process.

\subsubsection{loss function for binary label}
We mainly adapt the cross entropy loss for binary classification to form this loss function. To deal the highly imbalanced distribution of label, we introduce into the loss function a large weight factor $\alpha$ for label 1:
\begin{equation*}
loss(y^\star,\hat{y})=\sum_{i}^{}(-\alpha\ y_i^\star\log(\hat{y_{i,1}})-(1-y_i^\star)\log(1-\hat{y_{i,0}}))
\end{equation*}

\subsubsection{loss function for exponential-decay label}
While the mean square error loss is a natural choice for the exponential-decay label, we find during experiment that the output of the LSTM variants trained with naive MSE loss is smooth around the peak, because it's not sensitive to the maximum value. Let's say the initial output is 0.5 everywhere in the sequence, the loss measured by MSE is almost the same on the boundaries and in the middle of the music sequence. But what would be preferable is to distinguish the output score on the boundary from its neighbor, since this would allow us to raise the maximum picking threshold and acquire higher accuracy. Therefore we add a penalty mechanism in order to better fit the exponentially decay labels. Inspired by focal loss \cite{focalloss}, we introduce an additional rescaling factor into our MSE loss function for exponential-decay label:
\begin{equation*}
loss(y^\star, \hat{y}) = \frac{1}{n} \ \sum{z_i}
\end{equation*}
\begin{equation*}
z_i =\begin{cases}
\alpha \ \text{${(\hat{y_i} - y_i^\star)}^2$} & \text{if $y_i^\star = 1 \ and \ |\hat{y_i} - y_i^\star|< 1$}\\
\text{$\frac{1}{2} \ {(\hat{y_i} - y_i^\star)}^2$} & \text{if $y_i^\star \neq 1 and \ |\hat{y_i} - y_i^\star|< 1$}\\
\text{$|\hat{y_i} - y_i^\star| - \frac{1}{2}$} & \text{otherwise}
\end{cases}       
\end{equation*}

\subsubsection{loss function for linear-ascend label}

Since only the training of CRF variants involves the use of linear-ascend label, we present here the standard loss function of CRF that seeks to maximize the negative log-likelihood of the ground-truth sequence label:

\begin{align*}
loss(x, y^\star) = \sum_{\hat{y}} \sum_{i=0}^{n} Log(P(x_i|\hat{y_i})T(\hat{y_i}|\hat{y_{i-1}}))\\
-\sum_{i=1}^{n} Log(P(x_i|y_i^\star)T(y_i^\star|y_{i-1}^\star))
\end{align*}

Note that during training, CRF does not need to produce any prediction sequence $\hat{y}$. Instead, it works with all possible label sequences given $x$.

\section{EXPERIMENTS} \label{sec:experiments}

\subsection{Dataset}
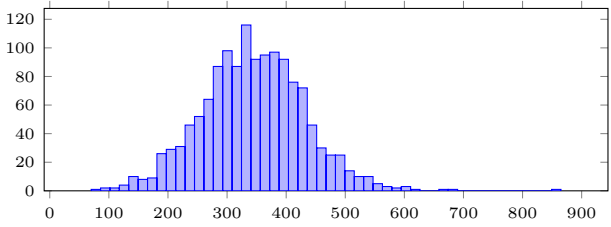
\begin{figure}[t]
\pgfplotsset{compat=1.14}
\label{song_length}
\begin{tikzpicture}
\begin{axis}[
	tiny,
    height=4cm,
    width=9cm,
    xtick={0,100,200,300,400,500,600,700,800,900},
    ymin=0,
    ybar
]
\addplot +[
    hist={
        bins=50
    }   
] table [y index=0] {song.csv};
\end{axis}
\end{tikzpicture}
\caption{distribution of song lengths over the dataset}
\end{figure}

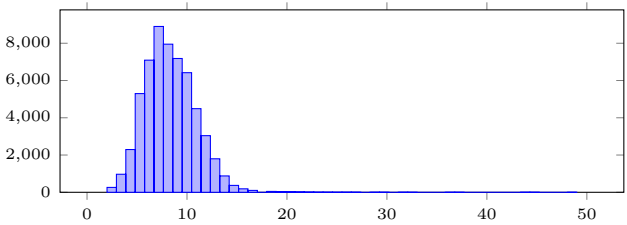
\begin{figure}[t]
\label{phrase_length}
\begin{tikzpicture}
\begin{axis}[	
	tiny,
    height=4cm,
    width=9cm,
    xtick={0,10,20,30,40,50},
    ymin=0,
    ybar
]
\addplot +[
    hist={
        bins=50
    }   
] table [y index=0] {phrase.csv};
\end{axis}
\end{tikzpicture}
\caption{distribution of phrase lengths over the dataset}
\end{figure}

We collected about 1000 well-known Chinese pop songs with melody and phrase label in it. The longest song contains 865 notes and the shortest song contains 70 notes. The distribution of song length is shown in Figure~\ref{song_length}. After aligning the symbolic representations of each song with the corresponding phrase boundary labels, we obtain a dataset whose total number of music phrase is around 50,000. The length of music phrases range from 2 notes to 49 notes, with a distribution shown in Figure~\ref{phrase_length}. Most phrases have a length around 9 notes. The labels also conform with this pattern: with 1 representing end-of-phrase and 0 otherwise, the data set contains 88\% 0s and 12\% 1s, meaning that there indeed exists a label sparsity problem. Among all 1 labels, there are 9.5\% of them are special in that they have no rest or break between them and the next notes, so the model might have difficulty identifying this kind of phrase boundary.

\subsubsection{Data Augmentation}
Previous works have suggested that data augmentation can improve model's performance\cite{data_augmentation}. Hence, We also employed it during the training process. There are three types of transformations that we performed. Given a sequence of notes with representation form (pitch, duration, silence), first we may add an integer between 0 and 12 to the pitch dimension, then a non-negative constant to the duration dimension, and third another non-negative constant to the silence dimension, which produces an augmented training set at least 48x larger than the original one. There are other valid transformations that one can perform; for example, a rescaling on the duration and silence dimensions, but we did not include them in our works, for the size of the dataset after augmentation has already been large enough. 
\subsection{Training Settings}

Our implementation of all proposed architecture is based on Pytorch library \cite{pytorch}. Mini-batch stochastic gradient descent was used as optimization algorithm for all our architectures. Detailed settings like learning rates and loss functions can be found in the corresponding sub-sections.  One special note, however, for the LSTM and CRF variants: due to the exploding and vanishing gradient problem \cite{gradient}, these models cannot learn if we simply input the whole songs. Hence, during training, we chop songs further into some sequences of notes that contain complete phrases only, i.e., no phrase will be broken into two piece and shared by two adjacent sequences. For LSTM variants, each sequence of notes contains 5 phrases precisely. For CRF variants, each sequence of notes contains at least 80 notes and at least two, but an unknown number of complete phrases. All sequences have less than 120 notes, the upper-bound to which we pad our sequence length. We input the whole song only at validation time.

\begin{figure*}[t]
	\includegraphics[width=\linewidth]{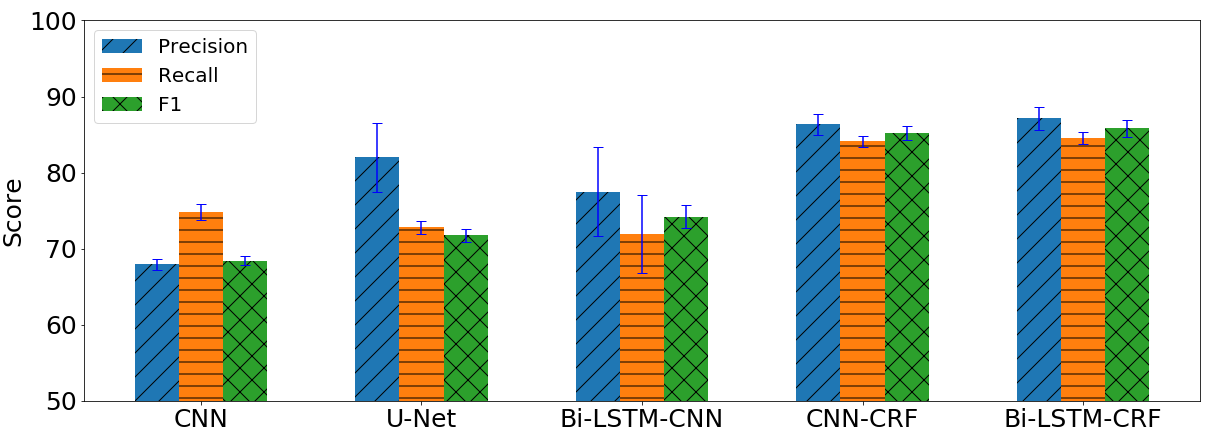}
    \caption{Best performance score by architectures}
    \label{best_figure}
\end{figure*}

\subsubsection{CNN Variants Settings}
Our simple 5-layer CNN uses 3,3,3,3,5 as the kernel size on each level. We trained our CNN models using mini-batch stochastic gradient descent \cite{minibatch} with batch size 100. In order to make the length for each song to be the same for mini-batches, we pad the original songs to a fixed length by repeating it from its beginning until it extends to the settled length. We set the length to 880, which is slightly bigger than the longest song. The loss used for each labeled data is mentioned in 4.4. For binary label, we find that the weight factor $\alpha=2$ performed best for our task after a grid search.

\subsubsection{LSTM Variants Settings}
As mentioned previously, we implement a 7-layer LSTM architecture with a 3-layer CNN feature extractor. The CNN feature extractor extrapolates the feature size to 32 dimensions and our LSTM units is initialized with hidden size 256. In addition, skip-connections \cite{skip-connection} are added among layers. In the training time, the Bi-LSTM-CNN model takes a mini batch of sequential input with length 100, which contains 5 complete music phrase before the zero padding. The loss is measured by the criterion mentioned in 5.4.2. Parameter optimization is performed with mini-batch SGD with batch size 32 and momentum 0.9. We choose learning rate of $\eta = 0.01$, and it is update after each epoch with decay rate $\varrho = 0.1$.  To prevent potential "gradient exploding" resulted from our penalty mechanism, we use a gradient clipping of 10.0 \cite{clipping}. 

The model trained in this way shows better capability of learning the peak value of the exponentially decay labels.  We use 5-fold cross validation to track the performance of our model while training. The validation loss goes up right after the first epoch. And in terms of accuracy, the precision rate stays high but the recall rate goes down rapidly. Therefore, several techniques are used to prevent Bi-LSTM-CNN model from over-fitting. Dropout layers are added right after the LSTM units and before the skip connection. L2 regularization and data augmentation are also tested, the result is shown in section 5.3. Note that in section 5.3, we choose Bi-LSTM-CNN model to represent LSTM variants. Dropping the CNN feature extractor will result in an approximately decline of $2\%$ in F1 score. This information is provided for those who are interested in the performance boost contributed by the CNN feature extractors and those who want to make comparison between Bi-LSTM and Bi-LSTM-CRF.

\subsubsection{CRF Variants Settings}
For CRF variants, the best learning rate and L2-regularization weight pair is 0.01 and 5e-8, with batch size 256. We also include a scheduler that scales down the learning rate by 0.75 once an epoch of training completes. Currently this is the best-performing setting we get from a grid search on the log space of hyper-parameters. We use the loss function in section 4.4.2 to train our CRF variants. One special setting when we adopt linear-ascend label as described in section 3.2 to train the CRF variants, however. When initializing the network, we put a large negative number at $A_{i,j}$, where $A$ is the transition matrix of the final CRF layer, if it is illegal for a sequence to jump from label $j$ to label $i$. This can enforce the network to take into consideration only prediction paths that are valid. 

\begin{figure*}
	\includegraphics[width=\linewidth]{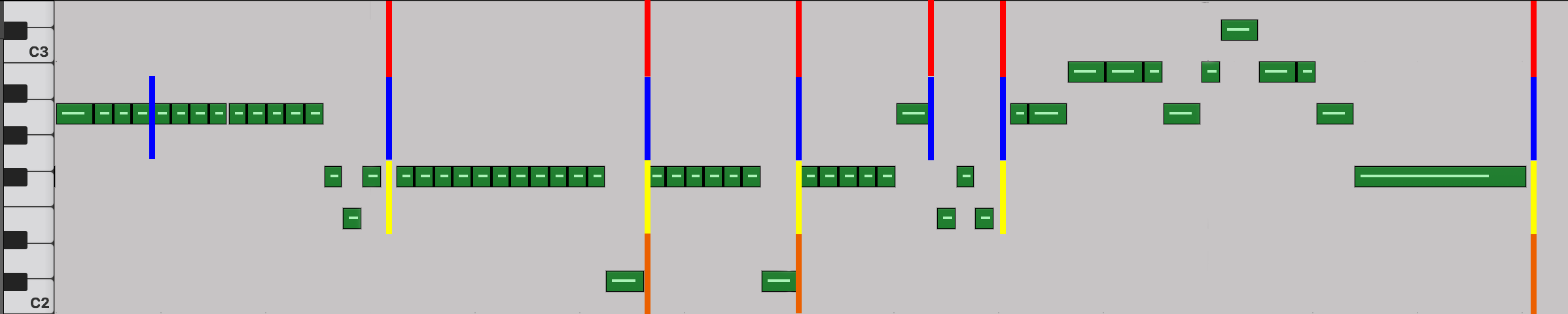}
    \caption{Cross Model Performance Comparison}
    \label{all}
\end{figure*}

\begin{figure*}
	\includegraphics[width=\linewidth]{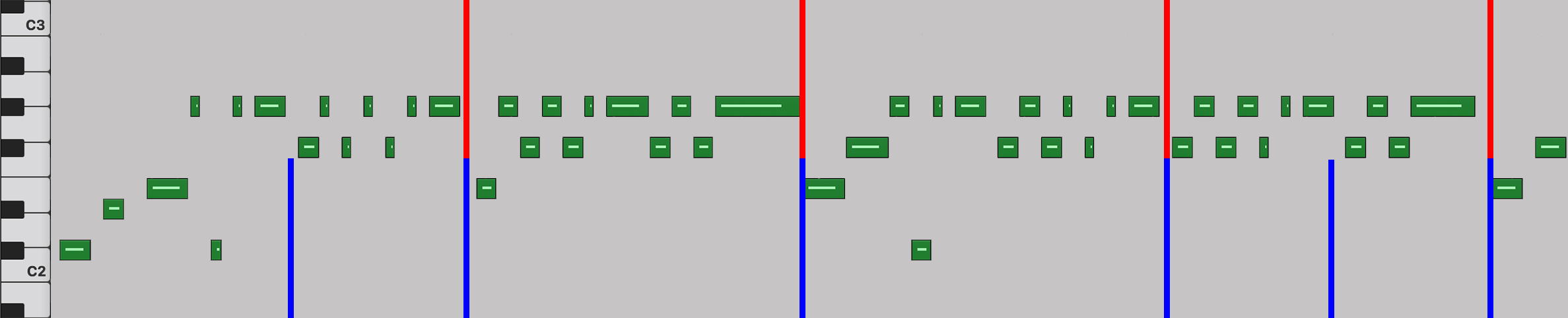}
    \caption{Phrase Boundary Prediction by CNN-CRF}
    \label{unique}
\end{figure*}

\begin{table}[b]
\addtolength{\tabcolsep}{-3pt}
\vskip 0.15in
\begin{center}
\begin{small}
\begin{sc}
\begin{tabular}{lcccr}
\toprule
Model & Precision & Recall & F1 \\
\midrule
CNN            & 67.94$\pm$ 0.72  & 74.82$\pm$ 1.08  & 68.44$\pm$ 0.580 \\
U-net\footnotemark  & 82.00$\pm$ 4.48 & 72.83$\pm$ 0.83 & 71.75$\pm$ 0.80 \\
Bi-LSTM-CNN    & 77.48$\pm$ 5.85  & 71.93$\pm$ 5.16  & 74.2$\pm$ 1.50   \\
CNN-CRF        & \textbf{86.37$\pm$ 1.38} & \textbf{84.11$\pm$ 0.70} & \textbf{85.22$\pm$ 0.96} \\
Bi-LSTM-CRF    & \textbf{87.13$\pm$ 1.51} & \textbf{84.59$\pm$ 0.76} & \textbf{85.83$\pm$ 1.10}\\

\bottomrule
\end{tabular}
\end{sc}
\end{small}
\end{center}
\vskip -0.1in
\caption{Model best performance score over all configurations}
\label{best}
\end{table}
\footnotetext{Because of the large standard deviation U-net has for the precision score, the averaged F1 score for U-net is lower than its precision and recall scores.}

\subsection{Results}
Experiment results show that the CRF variants are most suited for automated symbolic phrase segmentation task. Table~\ref{best} and Figure~\ref{best_figure} sums up the best performance each model can achieve over all different data label and augmentation configurations, along with their detailed precision and recall scores, averaged over 5-fold cross-validation. All models can achieve satisfactory F1 scores under certain settings; however, the CRF variants leads by a substantial margin. Further visualization of each model's prediction agrees with the performance ranking in Table~\ref{best}. For brevity, we include in Figure~\ref{all} only the corresponding ground-truth label of phrase boundaries and predictions given by the CNN, Bi-LSTM-CNN and CNN-CRF models on a music sample from the held-out test set, with red lines indicating there are phrase boundaries at these position according to the ground-truth label, and blue, yellow and orange lines indicating phrase boundary predictions by CNN-CRF, Bi-LSTM-CNN and CNN models respectively. The predictions given by Bi-LSTM-CRF and U-net are not included for they are the same as those given by CNN-CRF and CNN. As shown in Figure~\ref{all}, CNN model is only able to find half of the phrase boundaries, while Bi-LSTM-CNN model manages to find most of the phrase boundaries, missing only one place. CNN-CRF model not only successfully identifies all phrase boundaries according to the ground-truth label, but also marks one more place as possible phrase boundary point. 

Another observation can be noted from Table~\ref{best} is that time series models, once properly set up, can perform much better than the CNN variants for this particular task. This can be explained by time series models' ability to capture long-term dependencies in theory, yet the performance boost comes at the expense of increased training time. Time series models typically take longer time to train under the settings described in section 5.2. Bi-LSTM-CNN and Bi-LSTM-CRF models take the longest time to train. While CNN-CRF trains much faster than Bi-LSTM-CRF and Bi-LSTM-CNN, its training is still slower than the training of CNN and U-net. Since CNN-CRF and Bi-LSTM-CRF perform roughly at the same level, we favor CNN-CRF for its much shorter training time.

Experiment results also shows that our label engineering techniques are indeed effective. Table~\ref{augmentation} and Table~\ref{without_augmentation} shows the best F1 score each model can achieve under different types of labeling schemes with augmentation and without augmentation,  where a cross at $i^{th}$ row, $j^{th}$ column implies that the $j^{th}$ labeling scheme is not applicable to the $i^{th}$ model. By comparing columns within Table~\ref{augmentation} and Table~\ref{without_augmentation}, it can be seen that applying label engineering can help improving neural networks' performance. For Bi-LSTM-CNN model, the use of exponential-decay label is so essential for the training process that without it, the F1 score deteriorates to 3.84 only. For CRF variants, although the performance boost does not seem to be very large quantitatively, the use of linear-ascend label generally makes the training of CRF variants more robust. While the training of CRF variants using binary label failed to find a satisfactory solution on many sets of hyper-parameters we searched, training of CRF variants using linear-ascend label can often produce a good solution with less constraints on hyper-parameters. 

\begin{table}[t]
\addtolength{\tabcolsep}{-3pt}
\vskip 0.15in
\begin{center}
\begin{small}
\begin{sc}
\begin{tabular}{lcccr}
\toprule
Model & Binary & Smooth & Ascending \\
\midrule
CNN          & 68.44$\pm$ 0.58 & 59.07$\pm$ 0.95 & $\times$       \\
U-net        & 71.75$\pm$ 0.80 & 70.01$\pm$ 0.50 & $\times$      \\
Bi-LSTM-CNN  & \textbf{3.84} $\pm$ 1.77 & 73.7 $\pm$ 1.4  & $\times$       \\
CNN-CRF      & 84.95$\pm$ 0.84 & $\times$        & 85.22$\pm$ 0.96\\
Bi-LSTM-CRF  & 84.97$\pm$ 0.95 & $\times$        & 85.83$\pm$ 1.10\\

\bottomrule

\end{tabular}
\end{sc}
\end{small}
\end{center}
\vskip -0.1in
\caption{Model averaged F1 score under various types of labeling systems with augmentation}
\label{augmentation}
\end{table}

\begin{table}[t]
\addtolength{\tabcolsep}{-3pt}
\vskip 0.15in
\begin{center}
\begin{small}
\begin{sc}
\begin{tabular}{lcccr}
\toprule
Model & Binary & Smooth & Ascending \\
\midrule
CNN          & 68.06$\pm$ 0.72          & 58.28$\pm$ 0.86  & $\times$     \\
U-net        & 70.82$\pm$ 0.69          & 62.27$\pm$ 0.66  & $\times$     \\
Bi-LSTM-CNN  & 3.57 $\pm$ 0.87          & 74.2$\pm$1.5     & $\times$       \\
CNN-CRF      & \textbf{75.93}$\pm$ 0.74          & $\times$         & 81.56$\pm$0.78 \\
Bi-LSTM-CRF  & \textbf{4.62}$\pm$ 0.05  & $\times$         & \textbf{14.28}$\pm$0.34 \\
\bottomrule
\end{tabular}
\end{sc}
\end{small}
\end{center}
\vskip -0.1in
\caption{Model averaged F1 score under various types of labeling systems without augmentation}
\label{without_augmentation}
\end{table}

There is a significant performance gap that worth noting as well between CRF variant with data augmentation and without augmentation in Table~\ref{augmentation} and Table~\ref{without_augmentation}. Without data augmentation, the F1 score of both CRF variants drop by a substantial amount, showing that the data augmentation process is an essential part to the training of CRF variants for this task. The use of linear-ascend label can to some extent mitigate the performance drop. The performance of CNN variants and Bi-LSTM-CNN model, on the other hand, does not seem to depend much on our data augmentation method. One small improvement of data augmentation to Bi-LSTM-CNN model is that it prevents over-fitting. The performance no longer drops right after the first epoch, and the results are steady across all 5-fold validation. 

In addition, our best-performing models are able to develop its own understanding of music. Figure~\ref{unique} visualizes the phrase boundaries the CNN-CRF model predicts on another music sample from our test set and the corresponding ground-truth label, following the same notation used in Figure~\ref{all}. There are places that can be deemed as the start or the end of a phrase, but the ground truth label chooses not to mark these places as boundaries. Our CNN-CRF model can not only successfully predict where the ground truth labels think a phrase begins or ends, but also identify these places as boundaries and produce a finer phrase segmentation. This finding confirms that our method performs well for the automated symbolic music phrase segmentation task.


\section{Conclusion} \label{sec:conclusion}
We introduce in this paper a set of deep learning architectures and two label engineering techniques for the symbolic music phrase segmentation task. Experiment results indicates the effectiveness of our label engineering techniques. While all models can yield satisfactory phrase segmentation, combining CRF with deep neural networks dramatically improves the performance of our models, as CRF explicitly characterizes the relation among labels. Considering both effectiveness and efficiency, CNN-CRF is favored for this specific task. Future work will involve an optimization on the network architectures and the application of the results to automated composition.

\bibliography{melodicPhraseSegmentation}

%
%
%
%


\end{document}